\documentclass{article} 
\usepackage{iclr2026_conference,times}

\usepackage{amsmath,amsfonts,bm}

\def\eqref#1{equation~\ref{#1}}

\def\1{\bm{1}}

\DeclareMathAlphabet{\mathsfit}{\encodingdefault}{\sfdefault}{m}{sl}
\SetMathAlphabet{\mathsfit}{bold}{\encodingdefault}{\sfdefault}{bx}{n}

\usepackage[utf8]{inputenc} 
\usepackage[T1]{fontenc}    
\usepackage{amsfonts}       
\usepackage{nicefrac}       
\usepackage{microtype}      
\usepackage{makecell}

\usepackage{hyperref}
\usepackage{url}
\usepackage{graphicx}
\usepackage{multirow}
\usepackage[table,xcdraw]{xcolor}
\usepackage{amsmath}
\usepackage{booktabs}
\usepackage[capitalize,noabbrev]{cleveref}
\usepackage{colortbl}
\usepackage{pifont}
\usepackage[normalem]{ulem}
\usepackage{wrapfig}
\usepackage[many]{tcolorbox}
\usepackage{listings}
\usepackage{algorithm}
\usepackage{algorithmic}
\usepackage{caption}

\title{Resolving Sentiment Discrepancy for Multimodal Sentiment Detection via Semantics Completion and Decomposition}

\author{
    \vspace{-0.8cm}\\
    \textbf{Daiqing Wu}$^{1,3}$\quad
    \textbf{Dongbao Yang}$^{1,*}$\quad
    \textbf{Huawen Shen}$^{1}$\quad
    \textbf{Can Ma}$^{1}$\quad
    \textbf{Yu Zhou}$^{2,*}$\\
    \vspace{-0.3cm}\\
    \footnotesize{$^1$IIE, Chinese Academy of Sciences}\quad
    \footnotesize{$^2$Nankai University}\quad
    \footnotesize{$^3$University of Chinese Academy of Sciences}\\
    \footnotesize{$^*$Corrsponding Aurthors} \vspace{0.1cm}\\
    \texttt{wudaiqing@iie.ac.cn}\,\,\,\, 
    \texttt{yzhou@nankai.edu.cn} \\
    \vspace{-0.5cm}
}

\iclrfinalcopy 
\begin{document}

\maketitle

\begin{abstract}
With the proliferation of social media posts in recent years, the need to detect sentiments in multimodal (image-text) content has grown rapidly. Since posts are user-generated, the image and text from the same post can express different or even contradictory sentiments, leading to potential \textbf{sentiment discrepancy}. However, existing works mainly adopt a single-branch fusion structure that primarily captures the consistent sentiment between image and text. The ignorance or implicit modeling of discrepant sentiment results in compromised unimodal encoding and limited performance. In this paper, we propose a semantics \underline{Co}mpletion and \underline{De}composition (\textbf{CoDe}) network to resolve the above issue. In the semantics completion module, we complement image and text representations with the semantics of the in-image text, helping bridge the sentiment gap. In the semantics decomposition module, we decompose image and text representations with exclusive projection and contrastive learning, thereby explicitly capturing the discrepant sentiment between modalities. Finally, we fuse image and text representations by cross-attention and combine them with the learned discrepant sentiment for final classification. Extensive experiments on four datasets demonstrate the superiority of CoDe and the effectiveness of each proposed module.
\end{abstract}

\section{Introduction}
\label{sec:intro}

Multimodal sentiment detection aims to detect sentiments embedded in image-text posts. With the popularity of social media and the exponential increase in posts, multimodal sentiment detection demonstrates broad applications in opinion mining, psychological health, and business intelligence \citep{tpami2021VSAreview}. Unlike unimodal data, image-text posts contain abundant information from both visual and textual modalities. The complexity of effectively encoding and fusing modalities becomes the core focus of researchers.

\begin{figure}[t]
	\centering
	\includegraphics[width=1\textwidth]{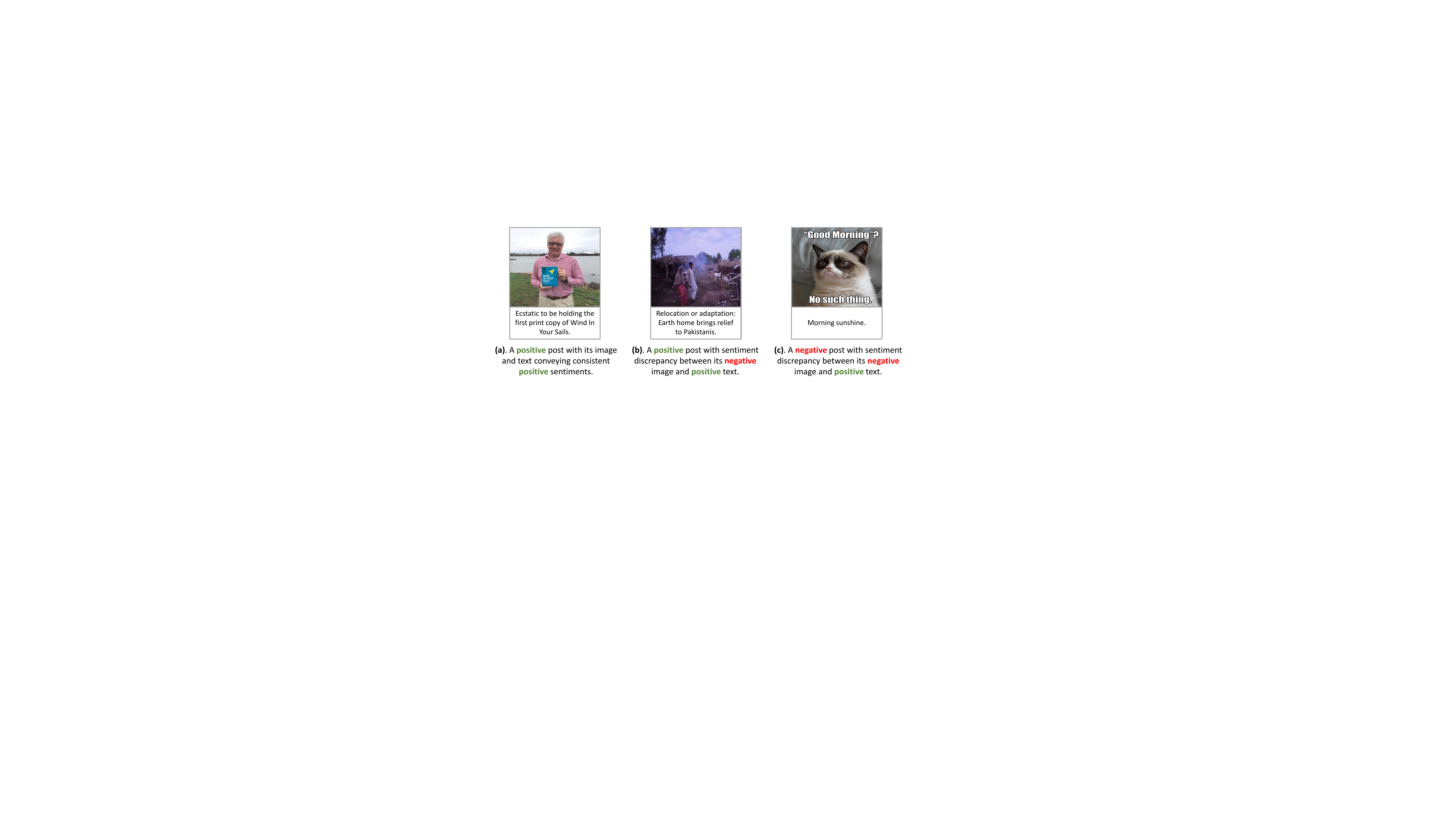}
    \vskip -0.1in
	\caption{Posts from social media.}
    \vskip -0.1in
	\label{demonstrated sample}
\end{figure}

Previous studies primarily focus on modeling inter-modal interactions \citep{yang2020mvan} and inter-sample relationships \citep{2023aclMVCN}. Despite their successes, they have not properly handled \textbf{sentiment discrepancy}\textemdash that the sentiments embedded in the image and text from the same post can be different or even contradictory, a prevalent issue in posts collected from social media. As illustrated in \cref{demonstrated sample} (b), the image portrays a scene of ruins after a disaster, evoking a negative sentiment. In contrast, the text describes an organization providing solutions for the refugees, shifting the overall sentiment to a positive tone. Compared to posts with consistent sentiments (like \cref{demonstrated sample} (a)), those with sentiment discrepancy pose a greater challenge for previous models. 
Proper handling of sentiment discrepancy also offers potential advantages in practical applications \citep{yue2019survey, emnlp18msd}. For instance, in user behavior analysis, it empowers brands to detect conflicting sentiments in user feedback, enabling timely intervention to prevent reputation risks. In political opinion mining, identifying sentiment discrepancy in politicians' tweets could separate sarcasm from the underlying intentions.

\begin{figure}[h]
	\centering
	\includegraphics[width=0.8\textwidth]{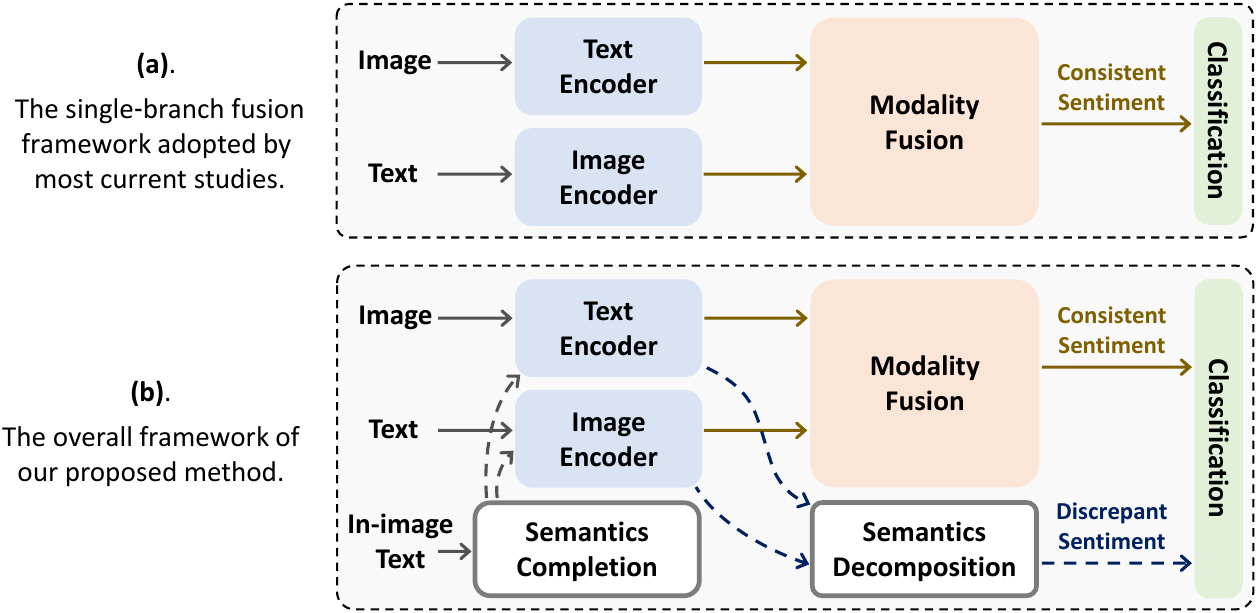}
    \vskip -0.1in
	\caption{Framework comparison of single-branch fusion and our method.}
    \vskip -0.1in
	\label{structure}
\end{figure}

Concretely, most current studies \citep{2023aclMVCN, pr2023msa} employ a single-branch fusion framework, as illustrated in \cref{structure} (a). Unimodal features are extracted separately and fed into a fusion module. Guided by the multimodal sentiment label, this framework encourages each unimodal encoder to capture the sentiment consistent with the multimodal label, which we refer to as the consistent sentiment. In contrast, the discrepant sentiment, embedded in unimodal semantics that is distinct from the multimodal label, is ignored or modeled implicitly. When encountering sentiment discrepancy during training\textemdash which accounts for 26.0\%-73.9\% (\cref{statistics}) of posts across datasets \citep{niu2016mvsa, cai2019hfm, yang2020mvan}, the consistent sentiment is insufficient to represent the entire post. Consequently, both image and text representations overfit to the consistent sentiment, compromising the unimodal encoding and, in turn, the fused multimodal representation. 

These limitations stem from the assumption within the design of the single-branch fusion framework that sentiments are consistent across modalities. However, this can be frequently violated in real-world scenarios, necessitating the explicit modeling of the discrepant sentiments between modalities. With this aim, we propose a semantics \underline{Co}mpletion and \underline{De}composition method, abbreviated as CoDe. As depicted in \cref{structure} (b), CoDe possesses two additional modules: semantics completion and semantics decomposition. In the semantics completion module, CoDe leverages the texts embedded in images (later referred to as in-image text) to complement both image and text representations semantically. According to statistics (\cref{statistics}), in-image text is present in the images of 63\% posts. It can furnish contextual semantics for both modalities, playing a crucial role in alleviating sentiment discrepancy. As shown in \cref{demonstrated sample} (c), the image depicts a grumpy cat with a negative sentiment, while the text conveys a positive attitude toward the morning. There exists an apparent sentiment discrepancy between them. The in-image text: ``Good Morning? No Such thing.'' can function as a semantic bridge, contributing to narrowing the sentiment gap.

In the semantics decomposition module, CoDe decomposes image and text representations into sub-representations, explicitly guiding them to learn modality-shared and modality-private sentiments. Specifically, modality-shared sentiments are semantically similar across modalities, constituting the consistent sentiment, and modality-private sentiments are unique to individual modalities, potentially leading to the discrepant sentiment. 

By leveraging these decomposed sub-representations, CoDe simultaneously models both discrepant and consistent sentiments, facilitating comprehensive encoding of the multimodal content. Through systematic evaluation, we demonstrate the effectiveness of CoDe's constituent modules, validating the benefits of resolving sentiment discrepancy. In summary, our contributions can be summarized as follows:

\begin{itemize}
    \item We propose CoDe to resolve sentiment discrepancy within image-text posts. It includes two modules in addition to the single-branch fusion framework: a semantics completion module that complements unimodal representations with in-image text semantics to alleviate sentiment discrepancy, and a semantics decomposition module that decomposes unimodal representations to explicitly capture the discrepant sentiment between modalities.
    \item We conduct extensive experiments on four multimodal datasets, demonstrating CoDe's superior performance over state-of-the-art methods. Furthermore, comprehensive ablation studies confirm the efficacy of our proposed modules. As the first attempt to handle sentiment discrepancy with in-image text, the semantics completion module effectively mitigates the sentiment gap between modalities. By decoupling modality-shared and modality-private sentiments, the semantics decomposition module provides benefits in promoting the encoding of holistic multimodal representations.
\end{itemize}

\section{Related Works}

\subsection{Unimodal Sentiment Detection}
Early studies on sentiment detection primarily focus on unimodal settings, including vision, language, and audio. To facilitate sentiment perception in models, pioneering researchers concentrate on handcrafted features, such as unigram models, aesthetic properties, and adjective-noun pairs. With advancements in deep learning and pre-trained models, the research focus has shifted to learning-based features. For instance, Kim \citep{kim-2014-cnn} utilizes pre-trained knowledge for visual sentiment analysis. Allaert \textit{et al.} \citep{taffc2022fer1} propose a novel Local Motion Pattern feature to advance facial emotion recognition, and Benjemaa \textit{et al.} \citep{cbmi2024fer2} further extend it to facial emotion spotting. Leveraging large-scale training data, these models demonstrate enhanced sentiment perception capabilities across diverse tasks.

\subsection{Multimodal Sentiment Detection}
\label{related}
Benefiting from the success of unimodal pre-trained models, multimodal sentiment detection has witnessed significant progress in recent years \citep{icml2025icl}. To fuse unimodal representations, numerous attention-based modules \citep{pr2023attention} are proposed to comprehensively capture the synergy between modalities. Among them, MultiSentiNet \citep{xu2017multisentinet} devises a visual-guided attention model, which is later extended to a bi-directional hierarchical version by HSAN \citep{xu2017hsan}. CoMN \citep{xu2018comemory} introduces a stackable memory hop structure, and MIMN \citep{xu2019mimn} refines it to the aspect-level. Recent works also consider inter-sample relationships. MGNNS \citep{yang2021mgnns} applies GNN to model the global characteristics of samples, and CLMLF \citep{li2022clmlf} employs contrastive learning among samples. More recently, MVCN \citep{2023aclMVCN} shifts focus toward tackling the granularity gap between modalities. Despite the success, their ignorance of the sentiment discrepancy leads to compromised unimodal encoding and incomplete multimodal representation. This drives us to resolve such issues and facilitate a comprehensive sentiment encoding for multimodal posts.

\subsection{Modality Gap} 
In multimodal data, various modalities have different semantics and information granularity, contributing to the modality gap. In recent years, it has drawn widespread attention across various fields, such as vision-language pre-training \citep{pr2024vlp}, multimodal crisis event categorization \citep{pr2023crisisevent}, multimodal sarcasm detection \citep{cvpr2023dip}, and video-targeted sentiment analysis \citep{hazarika2020misa}.

For studies focusing on \textbf{inter-modal semantics differences}, in vision-language pre-training, ALBEF \citep{neurips21albef} introduces a self-distillation structure by providing pseudo-targets from a momentum model. In crisis event categorization, Abavisani \textit{et al.} \citep{cvpr20crisisevent1} present a cross-attention module to filter uninformative and misleading components from weak modalities. In multimodal sarcasm detection, DIP \citep{cvpr2023dip} leverages sample distribution to model the inter-modal semantic similarity. Among these studies, DIP is most similar to ours. However, it aims to determine whether there exists sentiment discrepancy, while our model needs to perceive the specific discrepant sentiment. 

For studies focusing on \textbf{information granularity}, in video-targeted sentiment analysis, MISA \citep{hazarika2020misa} projects modalities into modality-invariant and -specific subspaces to minimize information redundancy. Based on MISA, TAILOR \citep{aaai2022tailor} proposes an adversarial refinement module to strengthen the diversity of each modality. Compared with them, we focus on sentiment discrepancy, which originates from \textbf{inter-modal semantics differences}. Inspired by their success, we design the semantics decomposition module based on MISA, and take one step further to guide the sentiment learning of decomposed representations. This enables our model to explicitly capture the modality-private sentiment. 

\subsection{In-image Text} 
Images frequently contain in-image texts, which are vital for human comprehension of visual content. Previous studies on object detection and recognition have utilized such information, verifying its benefits in understanding images. Karaoglu \textit{et al.} \citep{karaoglu2016words} use recognized text as complements to visual cues for logo retrieval. TextPlace \citep{hong2019textplace} utilizes the invariance of in-image text to illumination changes for robust visual place recognition. ViSTA \citep{cheng2022vista} devises a unified framework for deep interaction between modalities. Compared with serving as object accessories in natural scenes, in-image texts in user-generated posts—such as emoji fonts or personal comments—often inject subjective attitudes toward the depicted content. As a result, they convey more subjective information, extending beyond the literal visual representation. Related multimodal approaches achieve improvements by incorporating in-image text into their models, yet they neglect its connections with other modalities and only utilize it as independent knowledge. In contrast, we adopt the semantics of in-image text as shared contextual information to handle the sentiment discrepancy between image and text.

\begin{figure*}[h]
  \centering
  \includegraphics[width=1\textwidth]{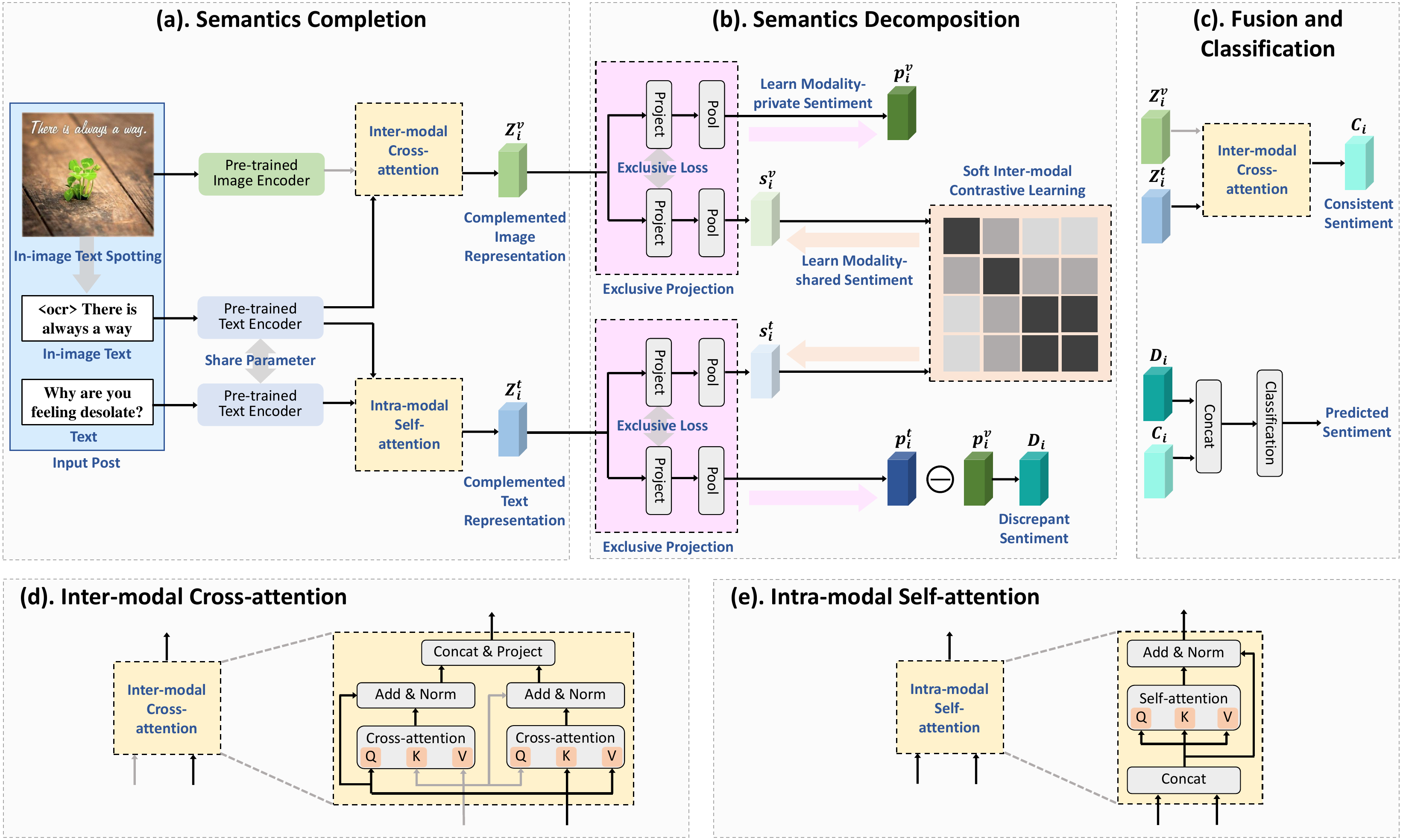}
  \vskip -0.1in
  \caption{Pipeline of CoDe. After encoding, two attention modules are developed to complement image and text representations with in-image text semantics. Following this, each complemented representation is decomposed into two components, with one learns the modality-shared sentiment, and the other learns the modality-private sentiment. Finally, the discrepant and consistent sentiments are explicitly modeled for classification.}
  \vskip -0.1in
  \label{fig3}
\end{figure*}

\section{Method}
\cref{fig3} demonstrates the overall architecture of Code, comprising the semantics composition module (\cref{fig3} (a)), the semantics decomposition module (\cref{fig3} (b)), the fusion and classification module (\cref{fig3} (c)), and the illustration for inter-modal cross-attention (\cref{fig3} (d)) and intra-modal self-attention (\cref{fig3} (e)).

\subsection{Task Definition}
\label{TD}
The objective of our model is to detect the specific sentiments embedded in image-text posts. The input is a set of posts:
\begin{equation}
\label{set of post}
\{(v_{1},t_{1}),(v_{2},t_{2}),\cdots,(v_{N},t_{N})\},
\end{equation}
where $v_{i}$ and $t_{i}$ represent the image and text of $i$ th post, respectively. The model needs to classify each post $(v_{i},t_{i})$ into a single sentiment category $l_{i}$ that belongs to a pre-defined category set.

\subsection{Semantics Completion Module}
\label{SCO}
As shown in \cref{fig3} (a), given a post $(v_{i},t_{i})$, we first process image $v_{i}$ by two off-the-shelf models, DBNet \citep{liao2020dbnet} and SAR \citep{li2019sar}, to extract in-image text $o_{i}$. DBNet detects the potential regions containing in-image text, and SAR recognizes the contents within these regions. $o_{i}$ is a sequence of words, similar to text $t_{i}$.

We then encode image $v_{i}$ by pre-trained image encoder SwinT \citep{liu2021swin}, text $t_{i}$ by pre-trained text encoder BERT \citep{jacob2019bert}, to obtain image representation $z_{i}^{v}\in \mathbb{R}^{I \times F}$ and text representation $z_{i}^{t}\in \mathbb{R}^{L \times F}$. $I$ is the number of image patches, $L$ is the max text length, and $F$ is the feature dimension. Since in-image text $o_{i}$ resides in the textual modality, we extract its semantics similarly to text $t_{i}$. Specifically, we add an \textless ocr\textgreater\, token at the beginning of $o_{i}$ and feed it into the same BERT encoder that processes $t_{i}$, obtaining its representation $z_{i}^{o}\in \mathbb{R}^{L \times F}$.

To integrate the semantics of in-image text into image and text representations, we adopt two typical attention formulations. As a preliminary, an attention module is generally formulated as:
\begin{equation}
\label{attention1}
A(q, k, v) = softmax(\frac{QK^{T}}{\sqrt{d}})V.
\end{equation}
\begin{equation}
\label{attention0}
Q = qP_Q, K = kP_K, V = vP_V.
\end{equation}
Inputs query($q$), key($k$), value($v$) are first linear projected to a common subspace by projectors $P_Q, P_K, P_V$. Then, the correlation between $Q$ and $K$ is calculated via dot product similarity divided by the square root of feature dimension d (equals to $F$ in our cases) and a softmax normalization. Finally, $V$ is multiplied by this correlational matrix for information fusion. 

To complement image representation, we employ an inter-modal cross-attention, which is also illustrated in \cref{fig3} (d):
\begin{equation}
\label{attention5}
Z_{i}^{v} = P_{ca}[A(z_{i}^{o},z_{i}^{v},z_{i}^{v})+z_{i}^{o}, A(z_{i}^{v},z_{i}^{o},z_{i}^{o})+z_{i}^{v}] \in \mathbb{R}^{I \times F}.
\end{equation}
In the calculation, we first compute an in-image-text-guided attention, where $z_{i}^{o}$ serves as query, $z_{i}^{v}$ serves as key and value, and an image-guided attention, where $z_{i}^{v}$ serves as query, $z_{i}^{o}$ serves as key and value. We then concatenate them (denoted by the [$\cdot$]) and feed the result into linear projector $P_{ca} \in \mathbb{R}^{I \times (I+L)}$. The obtained complemented image representation is denoted as $Z_{i}^{v}$.

To complement text representation, we employ an intra-modal self-attention, as also illustrated in \cref{fig3} (e):
\begin{equation}
\label{attention2}
Z_{i}^{t} = A([z_{i}^{t}, z_{i}^{o}], [z_{i}^{t}, z_{i}^{o}], [z_{i}^{t}, z_{i}^{o}]) + [z_{i}^{t}, z_{i}^{o}] \in \mathbb{R}^{2L \times F}.
\end{equation}
In the calculation, we first concatenate $z_{i}^{t}$ and $z_{i}^{o}$. The obtained $[z_{i}^{t}, z_{i}^{o}]$ is later used to compute attention by simultaneously serving as query, key, and value. The obtained complemented text representation is denoted as $Z_{i}^{t}$.

The asymmetry in formulations between $Z_{i}^{v}$ and $Z_{i}^{t}$ arises from the differences between modalities. Both text $t_{i}$ and in-image text $o_{i}$ reside in textual modality and are encoded by the same BERT encoder. In this encoding, each channel of $z_{i}^{t}$ and $z_{i}^{o}$ represents the semantics of a textual word, allowing for direct concatenation to maintain feature consistency across dimensions. However, in the case of image representation $z_{i}^{v}$, where each channel represents the semantics of a visual patch, such consistency no longer holds. Directly concatenating them would introduce additional alignment burdens during early training stages, which is not conducive to the model's learning process. 

Within this module, the attention mechanism effectively integrates the semantics of in-image text into both image and text representations. By providing shared contextual information, this process bridges the semantics gap between image and text, thereby mitigating the sentiment discrepancy.

\subsection{Semantics Decomposition Module}
\label{SDE}
After obtaining the complemented representations $Z_{i}^{v}$ and $Z_{i}^{t}$, we decompose their semantics by exclusive projection. By supervising one component to learn modality-shared sentiment through contrastive learning, the other component is compelled to embed modality-private sentiment that potentially leads to cross-model discrepant sentiment. This process is illustrated in \cref{fig3} (b). Specifically, the exclusive projection is formulated as:
\begin{equation}
\label{image decomposition}
s_{i}^{v} = Pool(Z_{i}^{v}P_{s}^{v}) ,\quad p_{i}^{v} = Pool(Z_{i}^{v}P_{p}^{v}) \in \mathbb{R}^{F}.
\end{equation}
\begin{equation}
\label{text decomposition}
s_{i}^{t} = Pool(Z_{i}^{t}P_{s}^{t}) ,\quad p_{i}^{t} = Pool(Z_{i}^{t}P_{p}^{t}) \in \mathbb{R}^{F}.
\end{equation}
We feed $Z_{i}^{v}$ and $Z_{i}^{t}$ into four projectors. Each projector is followed by an average pooling layer (denoted by Pool($\cdot$)), which aggregates token-level features by computing the mean over the sequence dimension. This yields modality-shared sub-representations $s_{i}^{v}, s_{i}^{t}$ and modality-private sub-representations $p_{i}^{v}, p_{i}^{t}$. Projectors $P_{s}^{v}$, $P_{p}^{v}$, $P_{s}^{t}$, and $P_{p}^{t}$ are learnable $\mathbb{R}^{F \times F}$ matrices with row-wise normalization. They are constrained by an exclusive loss, with a hyperparameter $\sigma$ controlling the distinction between them:
\begin{equation}
\label{diversity loss3}
\mathcal{L}_{exc} = max(\sigma - ||P_{s}^{v}-{P^{v}_{p}}||_{F}, 0) + max(\sigma - ||P_{s}^{t}-{P_{p}^{t}}||_{F}, 0).
\end{equation}

\begin{table}[t]
  \caption{Mapping from sentiment categories (left columns) to discrete numerical values (right columns) of datasets \citep{yang2020mvan, cai2019hfm, niu2016mvsa}. MVSA-* refers to MVSA-Single and MVSA-Multiple.}
  \vskip -0.1in
  \centering
  \resizebox{0.65\linewidth}{!}{
  \renewcommand\arraystretch{1.1}
  \begin{tabular}{cc|cccccc|cc}
    \toprule
    \multicolumn{2}{c|}{\textbf{MVSA-*}}    & \multicolumn{6}{c|}{\textbf{TumEmo}} & \multicolumn{2}{c}{\textbf{HFM}}\\
    Positive & 0                            & Love   & 0   & Happy  & 1  & Calm   & 2         & Positive & 0      \\ 
    Neutral  & 1                            & Bored  & 3   & Sad & 4  & Angry  & 5         & Negative & 1      \\
    Negative & 2                            & Fear   & 6   &     &    & &    \\
    \bottomrule 
  \end{tabular}}
  \vskip -0.1in
  \label{weight map}
\end{table}

\begin{figure}[t]
  \centering
  \includegraphics[width=0.85\linewidth]{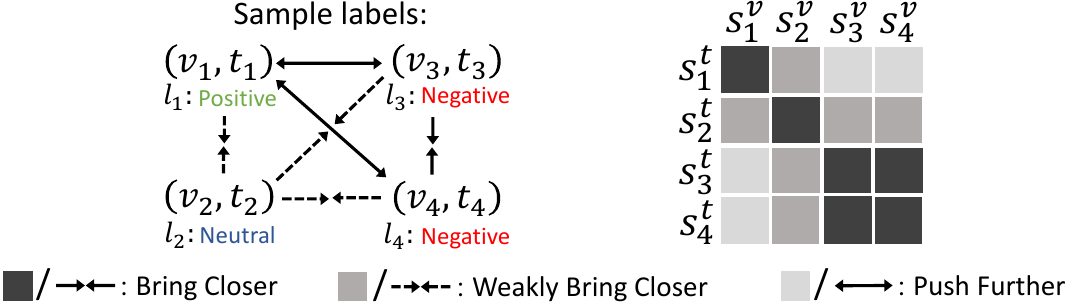}
  \vskip -0.1in
  \caption{An example of the soft inter-modal contrastive learning. Solid arrows pointing inward and black blocks indicate that the representations are brought closer. Dashed arrows pointing inward and grey blocks indicate the target of weakly bringing closer. The solid arrows pointing outward and the light grey blocks signify negative pairs to be pushed further. Each $s_{i}^{v}$ is brought closer to its counterpart $s_{i}^{t}$. $(s_{1}^{v},s_{3}^{t})$, with $w_{1,3}=0$, is treated as a negative pair and pushed away. $(s_{1}^{v},s_{2}^{t})$, with $w_{1,2}=0.5$, are treated as a partial positive pair and weakly brought closer. $(s_{3}^{v},s_{4}^{t})$, with $w_{3,4}=1$, is treated as a positive pair and brought closer.}
  \vskip -0.1in
  \label{fig4}
\end{figure}

Subsequently, we devise a soft inter-modal contrastive learning to supervise $s_{i}^{v}, s_{i}^{t}$ in learning the modality-shared sentiment. We construct each sample pair with a modality-shared image sub-representation and a modality-shared text sub-representation, and treat pairs with the same sentiment categories as positive pairs. Different from regular contrastive learning that treats all other pairs as negative pairs, we consider sample pairs with closely related but non-identical sentiment categories as partial positive pairs. The contrastive loss is thereby formulated as follows:
\begin{equation}
\label{Soft cross-modality contrastive loss1}
\mathcal{L}_{con} = -log (\sum_j w_{i,j} exp{(s_{i}^{v} \cdot s_{j}^{t}/\tau)}/\sum_j exp{(s_{i}^{v} \cdot s_{j}^{t}/\tau)}).
\end{equation}
$\tau$ is the temperature hyperparameter. Weight $w_{i,j}$ reflects the correlation between posts $(v_{i},t_{i})$ and $(v_{j},t_{j})$'s respective sentiment labels $l_i$ and $l_j$. In order to acquire such correlations, we construct a mapping $M$ (as shown in \cref{weight map}) that converts sentiment categories to discrete numerical values by referring to Ekman's emotion model \citep{ekman1992argument}. Following this, we define the weight $w_{i,j}$ based on the normalized distance between $M(l_i)$ and $M(l_j)$: 
\begin{equation}
\label{normalized distance}
w_{i,j} = 1- \frac{|M(l_i)-M(l_j)|}{|max(M(*))-min(M(*))|} \in [0,1].
\end{equation}
An example is illustrated in \cref{fig4} for an intuitive understanding.

Consequently, $s_{i}^{v}$ and $s_{i}^{t}$ learn the modality-shared sentiment. Constrained by the exclusive projection, $p_{i}^{v}$ and $p_{i}^{t}$ are compelled to encode semantics that are distinct from the modality-shared sentiment, namely, the modality-private sentiment. Since the modality-private sentiments are the primary source of sentiment discrepancy across modalities, we capture the discrepant sentiment $D_{i}$ by measuring the difference between $p_{i}^{v}$ and $p_{i}^{t}$, as formulated by:
\begin{equation}
\label{disparity}
D_{i} = p_{i}^{t} - p_{i}^{v} \in \mathbb{R}^{F}.
\end{equation}

\subsection{Fusion and Classification Module}
\label{fusion and classification}
Besides the discrepant sentiment, the consistent sentiment between image and text is also a crucial component of the comprehensive multimodal representation. Therefore, we capture it with a global fusion of $Z_{i}^{v}$ and $Z_{i}^{t}$ by employing the inter-modal cross-attention similar to \cref{attention5}:
\begin{equation}
\label{fusion}
C_{i} = Pool([A(Z_{i}^{v},Z_{i}^{t},Z_{i}^{t})+Z_{i}^{v}, A(Z_{i}^{t},Z_{i}^{v},Z_{i}^{v})+Z_{i}^{t}]) \in \mathbb{R}^{F}.
\end{equation}
$Pool(\cdot)$ represents an average pooling layer. Finally, we concatenate the consistent sentiment $C_{i}$ and the discrepant sentiment $D_{i}$, and feed the result into a projector $P_{cls}$ for classification:
\begin{equation}
\label{classification loss}
\mathcal{L}_{cls} = Cross-entropy([C_{i}, D_{i}]P_{cls}).
\end{equation}
The total loss is the combination of $\mathcal{L}_{cls}$, $\mathcal{L}_{exc}$ and $\mathcal{L}_{con}$:
\begin{equation}
\label{total loss}
\mathcal{L}_{total} = \mathcal{L}_{cls} + \mathcal{L}_{exc} + \mathcal{L}_{con}.
\end{equation}

\section{Experiment}

\subsection{Datasets}
We conduct experiments on four publicly available multimodal sentiment datasets that are prevalent for evaluating multimodal sentiment detection methods: MVSA-Single \citep{niu2016mvsa}, MVSA-Multiple \citep{niu2016mvsa}, TumEmo \citep{yang2020mvan}, and HFM \citep{cai2019hfm}. The detailed statistics are presented in \cref{statistics}. \textbf{MVSA-Single} is collected from Twitter\footnote{https://twitter.com}, with each post annotated with an image sentiment and a text sentiment from: \{Positive, Neutral, Negative\}. Following Xu \textit{et al.} \citep{xu2017multisentinet}, we assign multimodal categories to posts with majority voting. \textbf{MVSA-Multiple} is a larger version of MVSA-Single. \textbf{TumEmo} is a weakly-supervised dataset collected from Tumblr\footnote{http://tumblr.com}, where each post is labeled by the search keyword used to retrieve it. It contains 7 emotion categories: \{Angry, Bored, Calm, Fear, Happy, Love, Sad\}. \textbf{HFM} is a popular multimodal sarcasm detection dataset. The categories of HFM represent whether the post expresses sarcasm rather than the post sentiment. It comprises two categories: \{Positive, Negative\}. We follow the pre-processing approaches and adopt the evaluation metrics in \citep{yang2020mvan} and \citep{cai2019hfm}. Specifically, we report accuracy (\textbf{acc}) and weighted-F1 (\textbf{w-F1}) for MVSA-Single, MVSA-Multiple and TumEmo; accuracy (\textbf{acc}) and macro-F1 (\textbf{m-F1}) for HFM.

\begin{table}[t]
  \centering
    \caption{Statistics of datasets. Column 6 represents the number of posts containing in-image texts. Column 7 represents the number of posts containing sentiment discrepancy (SD). Column 8 represents the number of posts containing both ot them.}
    \vskip -0.1in
  \resizebox{1\linewidth}{!}{
  \begin{tabular}{l|c|ccc|ccc}
    \toprule
    \textbf{Dataset} & \textbf{Total}& \textbf{Train}& \textbf{Val} & \textbf{Test}&  \textbf{In-image Text (\%)} & \textbf{SD (\%)} & \textbf{In-image Text\&SD (\%)} \\
    \midrule
    MVSA-Single \citep{niu2016mvsa} & 4511 & 3608  & 451   & 452   & 2747 (60.9\%) & 1919 (42.5\%) & 1249 (27.7\%) \\
    MVSA-Multiple \citep{niu2016mvsa} & 17024 & 13618 & 1703  & 1703  & 11286 (66.3\%) & 4425 (26.0\%) & 3320 (19.5\%)\\
    TumEmo \citep{yang2020mvan} & 195265  & 156217 & 19524 & 19524   & 120491 (61.7\%) & 128907 (66.0\%) & 80679 (41.3\%)\\
    HFM \citep{cai2019hfm} & 24635 & 19816 & 2410 & 2409  & 18511 (75.1\%) & 18208 (73.9\%) & 13388 (54.3\%)\\
    \bottomrule
  \end{tabular}}
    \label{statistics}
    \vskip -0.1in
\end{table}

\subsection{Implementation Details}
All our implementations are based on PyTorch. We adopt the same settings across all four datasets. Our model is trained for 20 epochs with a batch size of 16. The initial learning rate is set to 5e-5 for the image encoder, 2e-5 for the text encoder, 1e-3 for the classification layer, and 2e-4 for other parameters.  We adopt AdamW optimizer and decay the learning rate using a cosine scheduler. Threshold $\sigma$ in \cref{diversity loss3} is set to 0.1, temperature $\tau$ in \cref{Soft cross-modality contrastive loss1} is set to 0.07.

\subsection{Baselines}
For unimodal comparison, we choose \textbf{CNN} \citep{kim-2014-cnn}, \textbf{Bi-LSTM} \citep{zhou2016bilstm}, \textbf{BERT} \citep{jacob2019bert} as text baselines, and \textbf{ResNet} \citep{he2016resnet}, \textbf{ViT} \citep{dosovitskiy2020vit}, \textbf{SwinT} \citep{liu2021swin} as image baselines. These models are commonly adopted for classification tasks in their corresponding modalities. 

For multimodal comparison, as introduced in \cref{related}, we choose \textbf{MultiSentiNet} \citep{xu2017multisentinet}, \textbf{HSAN} \citep{xu2017hsan}, \textbf{CoMN} \citep{xu2018comemory}, \textbf{MVAN} \citep{yang2020mvan}, \textbf{MGNNS} \citep{yang2021mgnns}, \textbf{CLMLF} \citep{li2022clmlf} and \textbf{MVCN} \citep{2023aclMVCN} as baselines. We notice a disagreement among these works on selecting unimodal encoders, which may introduce unfair comparisons. Therefore, we additionally reproduce two latest SOTA methods: \textbf{CLMLF} and \textbf{MVCN}, together with two classic fusion strategies: concatenation (\textbf{Concat}) and cross-attention (\textbf{Att}) based on the same SwinT and BERT encoders adopted by our model.

\begin{table}[t]
  \centering
      \caption{Comparison with SOTA unimodal and multimodal methods. In addition to the results reported in previous works, we report the results of four multimodal methods under the same unimodal encoders as ours for a fair comparison. $\ddagger$ indicates the reproductive operation. All reproduced results and our method are reported as the average over 10-fold cross-validation. \textit{Bold} indicates the best result. \textit{\underline{Underline}} indicates the second best result.}
      \vskip -0.1in
  \resizebox{1\linewidth}{!}{
  \begin{tabular}{c|l|cc|cc|cc|cc}
    \toprule
    \multirow{2}*{\textbf{Modality}} & \multirow{2}*{\textbf{Method}} & \multicolumn{2}{c|}{\textbf{MVSA-Sinlge}} & \multicolumn{2}{c|}{\textbf{MVSA-Multiple}}
      & \multicolumn{2}{c|}{\textbf{TumEmo}} & \multicolumn{2}{c}{\textbf{HFM}} \\
                                       &       & Acc     & W-F1     & Acc     & W-F1     & Acc   & W-F1     & Acc     & M-F1      \\          
    \midrule
    \multirow{3}*{Image}      & ResNet-50 \citep{he2016resnet} & 64.67  & 61.55 & 61.88  & 60.98  & 48.10  & 47.75  & 72.77  & 71.38\\
                                       & ViT \citep{dosovitskiy2020vit} & 63.78  & 62.26 & 61.94  & 61.19  & 46.35  & 45.94  & 73.09  & 71.52\\
                                       & $\text{SwinT}^\ddagger$ \citep{liu2021swin} & 64.89  & 63.94 & 62.44  & 61.09  & 49.60  & 48.27  & 72.82  & 72.53\\
                                       
    \midrule
    \multirow{3}*{Text}       & CNN \citep{kim-2014-cnn}       & 68.19  & 55.90 & 65.64  & 57.66  & 61.54  & 47.74  & 80.03  & 75.32\\
                                       & BiLSTM \citep{zhou2016bilstm}  & 70.12  & 65.06 & 67.90  & 67.90  & 61.88  & 51.26  & 81.90  & 77.53\\
                                       & BERT \citep{jacob2019bert}  & 71.11  & 69.70 & 67.59  & 66.24  & 62.12  & 60.78 & 83.89  & 83.26 \\

    \midrule                                
    \multirow{7}*{\makecell[c]{Image+Text \\ (Original Encoders)}}  & MultiSentiNet \citep{xu2017multisentinet} & 69.84  & 69.84 & 68.86  & 68.11 & -   & -      & -       & -       \\
                                       & HSAN \citep{xu2017hsan}  & 69.88  & 66.90 & 67.96  & 67.76  & 63.09  & 53.98 & -       & -      \\
                                       & CoMN \citep{xu2018comemory} & 70.51  & 70.01 & 69.92  & 69.83   & 64.26  & 59.09 & -       & -     \\
                                       & MVAN \citep{yang2020mvan}  & 72.98  & 71.39 & 71.83  & \underline{70.38}  & 65.53  & 65.43 & - & - \\
                                       & MGNNS \citep{yang2021mgnns} & 73.77  & 72.70 & \underline{72.49}  & 69.34  & 66.72  & 66.69 & - & - \\
                                       & CLMLF \citep{li2022clmlf} & 75.33  & 73.46 & 72.00  & 69.83  & - & - & 85.43  & 84.87\\
                                       & MVCN \citep{2023aclMVCN}  & \underline{76.06}  & 74.55 & 72.07  & 70.01  & -       & -    & 85.68  & 85.23   \\
    \midrule                                   
    \multirow{5}*{\makecell[c]{Image+Text \\ (SwinT+BERT)}} & $\text{Concat}^\ddagger$ & 72.99 & 72.21 & 69.63 & 68.53 & 70.89 & 70.16 & 84.93 & 84.22 \\
                                       & $\text{Att}^\ddagger$ & 74.61 & 74.11 & 70.07 & 68.17 & 71.01 & 70.53 & 84.78 & 84.75 \\
                                       & $\text{CLMLF}^\ddagger$ \citep{li2022clmlf} & 75.67 & 75.01 & 72.23 & 69.67 & 70.66 & 70.57 & 86.82 & 86.69\\ 
                                       & $\text{MVCN}^\ddagger$ \citep{2023aclMVCN} & 75.74 & \underline{75.64} & 72.09 & 69.58 & \underline{71.97} & \underline{71.37} & \underline{87.13} & \underline{86.76}\\ 
                                       & CoDe (\textbf{Ours}) & \textbf{76.98}  & \textbf{76.47} & \textbf{72.74}  & \textbf{70.71}  & \textbf{73.02}  & \textbf{73.02} & \textbf{88.88}  & \textbf{88.72} \\
    
    \bottomrule
  \end{tabular}}
  \vskip -0.1in
    \label{VTSA Result}
\end{table}

\begin{table}[t]
  \centering
      \caption{Student t-test results (p-values) between proposed method and reproduced baselines. All p-values are computed based on 10-fold cross-validation. \textit{Bold} (p \textless 0.05) highlights statistically significant improvements.}
      \vskip -0.1in
  \resizebox{1\linewidth}{!}{
  \begin{tabular}{l|cc|cc|cc|cc}
    \toprule
    \multirow{2}*{\textbf{Method}} & \multicolumn{2}{c|}{\textbf{MVSA-Sinlge}} & \multicolumn{2}{c|}{\textbf{MVSA-Multiple}}
      & \multicolumn{2}{c|}{\textbf{TumEmo}} & \multicolumn{2}{c}{\textbf{HFM}} \\
                                       & Acc     & W-F1     & Acc     & W-F1     & Acc   & W-F1     & Acc     & M-F1      \\          
    \midrule
    $\text{CLMLF}^\ddagger$ \citep{li2022clmlf} (SwinT+BERT) & \textbf{0.002}  & \textbf{\textless 0.001} & \textbf{0.045} & \textbf{\textless 0.001} & \textbf{\textless 0.001} & \textbf{\textless 0.001} & \textbf{\textless 0.001} & \textbf{\textless 0.001} \\
    $\text{MVCN}^\ddagger$ \citep{2023aclMVCN} (SwinT+BERT) & \textbf{0.005} & \textbf{0.033} & \textbf{0.012} & \textbf{\textless 0.001} & \textbf{\textless 0.001} & \textbf{\textless 0.001} & \textbf{\textless 0.001} & \textbf{\textless 0.001} \\
    
    \bottomrule
  \end{tabular}}
    \vskip -0.1in
    \label{student t-test}
\end{table}

\subsection{Comparison with SOTA methods}
We present the comparison results and statistical tests of CoDe and SOTA methods in \cref{VTSA Result} and \cref{student t-test}. Based on the results, we make the following observations: \textbf{1)}. CoDe consistently outperforms unimodal methods in all metrics, affirming the benefits of richer semantics in multimodal data for the comprehensive understanding of posts. \textbf{2)}. Compared with multimodal methods, CoDe also achieves significant performance gains across all metrics on four datasets. These results demonstrate the necessity of resolving sentiment discrepancy in detecting sentiment in posts. \textbf{3)}. CoDe achieves more obvious performance gains on TumEmo and HFM, while relatively modest gains on MVSA-Single and MVSA-Multiple. We attribute this to the category assignment strategy \citep{xu2017multisentinet} of MVSA-Single and MVSA-Multiple. It discards the posts with contradictory unimodal sentiment categories, thus significantly reducing the proportion of posts containing sentiment discrepancy, as presented in \cref{statistics}. For TumEmo, which is unfiltered and more representative of the real-world distribution of posts, and HFM, which contains frequent sarcastic expressions, the advantages of resolving sentiment discrepancy are much more highlighted.

\begin{table}
  \centering
    \caption{Ablation experiments of CoDe on MVSA-Single and TumEmo under two baselines. SCO abbreviates the semantics completion module. SDE abbreviates the semantics decomposition module. CL abbreviates contrastive learning. All our methods are reported as the average over 10-fold cross-validation.}
  \vskip -0.1in
  \resizebox{1\linewidth}{!}{
  \begin{tabular}{l|cc|cc|l|cc|cc}
    \toprule
    \textbf{Unimodal Baselines} & \multicolumn{2}{c|}{\textbf{MVSA-Sinlge}} & \multicolumn{2}{c|}{\textbf{TumEmo}} & \textbf{Unimodal Baselines} & \multicolumn{2}{c|}{\textbf{MVSA-Sinlge}} & \multicolumn{2}{c}{\textbf{TumEmo}} \\
    (SwinT+BERT)       & Acc     & W-F1     & Acc     & W-F1  & (ViT+BERT)       & Acc     & W-F1     & Acc     & W-F1\\              
    \midrule
    Image-only                      & 64.89  & 63.94 & 49.60 & 48.27 &
    Image-only                      & 63.78  & 62.26 & 46.35 & 45.94  \\
    Text-only                       & 71.11  & 69.70 & 62.12 & 60.78 & Text-only                       & 71.11  & 69.70 & 62.12 & 60.78 \\
    w/o SCO+SDE                     & 74.61  & 74.11 & 71.01 & 70.53  &
    w/o SCO+SDE                     & 74.20  & 73.27 & 64.12 & 63.98  \\
    w/o SCO                         & 75.39  & 74.15 & 72.49 & 72.49  &
    w/o SCO                         & 75.14  & 74.36 & 65.90 & 65.78  \\
    w/o Soft Inter-modal CL         & 76.01  & 75.59 & 72.21 & 72.16  &
    w/o Soft Inter-modal CL         & 76.08  & 75.30 & 67.29 & 67.20  \\
    w/o SDE                         & 76.15  & 75.78 & 72.21 & 72.17  &
    w/o SDE                         & 75.74  & 74.68 & 66.87 & 66.92  \\
    CoDe (\textbf{Ours})                           & \textbf{76.98}  & \textbf{76.47} & \textbf{73.02} & \textbf{73.02} & CoDe (\textbf{Ours}) & \textbf{76.32}  & \textbf{76.00} & \textbf{68.81} & \textbf{67.94}  \\
    \bottomrule
  \end{tabular}}
  \label{different modules}
  \vskip -0.1in
\end{table}

\begin{table}
  \centering
    \caption{Student t-test results (p-values) between ablated methods and the complete method. All p-values are computed based on 10-fold cross-validation. \textit{Bold} (p \textless 0.05) highlights statistically significant improvements.}
    \vskip -0.1in
  \resizebox{1\linewidth}{!}{
  \begin{tabular}{l|cc|cc|l|cc|cc}
    \toprule
    \textbf{Unimodal Baselines} & \multicolumn{2}{c|}{\textbf{MVSA-Sinlge}} & \multicolumn{2}{c|}{\textbf{TumEmo}} & \textbf{Unimodal Baselines} & \multicolumn{2}{c|}{\textbf{MVSA-Sinlge}} & \multicolumn{2}{c}{\textbf{TumEmo}} \\
    (SwinT+BERT)       & Acc     & W-F1     & Acc     & W-F1  & (ViT+BERT)       & Acc     & W-F1     & Acc     & W-F1\\              
    \midrule
    w/o SCO+SDE                     & \textbf{\textless 0.001}  & \textbf{\textless 0.001} & \textbf{\textless 0.001} & \textbf{\textless 0.001}  &
    w/o SCO+SDE                     & \textbf{\textless 0.001}  & \textbf{\textless 0.001} & \textbf{\textless 0.001} & \textbf{\textless 0.001}  \\
    w/o SCO                         & \textbf{\textless 0.001}  & \textbf{\textless 0.001} & 0.101 & \textbf{0.048}  &
    w/o SCO                         & \textbf{\textless 0.001}  & \textbf{\textless 0.001} & \textbf{\textless 0.001} & \textbf{\textless 0.001}  \\
    w/o Soft Inter-modal CL         & \textbf{0.001}  & \textbf{0.001} & \textbf{\textless 0.001} & \textbf{\textless 0.001}  &
    w/o Soft Inter-modal CL         & 0.140  & \textbf{0.001} & \textbf{\textless 0.001} & \textbf{\textless 0.001}  \\
    w/o SDE                         & \textbf{0.007}  & \textbf{0.025} & \textbf{\textless 0.001} & \textbf{\textless 0.001}  &
    w/o SDE                         & \textbf{\textless 0.001}  & \textbf{\textless 0.001} & \textbf{\textless 0.001} & \textbf{\textless 0.001}  \\
    \bottomrule
  \end{tabular}}
    \vskip -0.1in
  \label{ablation t-test}
\end{table}

\subsection{Ablation Experiments}
\label{ablation}
To probe the effectiveness of each module, we conduct ablation experiments and corresponding statistical tests in \cref{different modules} and \cref{ablation t-test} under two baselines. It can be observed under both baselines that: \textbf{1)}. Both the semantics completion module and the semantics decomposition module bring significant performance gains to the model in the majority of cases. This suggests that complementing image and text representations with the semantics of in-image text indeed alleviates the sentiment discrepancy, and explicitly capturing the discrepant sentiment leads to more comprehensive multimodal representations. \textbf{2)}. The removal of the soft inter-modal contrastive learning severely impacts the effectiveness of the semantics decomposition module under the SwinT+BERT baseline, underscoring the necessity of the sentiment guiding and feature alignment provided by the contrastive learning. \textbf{3)}. The semantics completion module improves the model stably on two datasets, while the semantics decomposition module is more efficient on TumEmo than on MVSA-Single. We conjecture that the contextual semantics of in-image text also benefits the model under unremarkable sentiment discrepancy, while the semantics decomposition module is more suitable for handling explicit sentiment discrepancy. This further explains the difference in performance improvements of CoDe on these datasets. \textbf{4)}. CoDe with both modules achieves the best performance, showing that the modules are complementary to each other.

\begin{table}[]
\centering
\caption{Comparison with advanced unimodal and multimodal methods on aspect-based multimodal sentiment analysis.}
\vskip -0.1in
\resizebox{0.7\linewidth}{!}{
\begin{tabular}{c|l|cc|cc}
\toprule
\multirow{2}{*}{\textbf{Modality}}   & \multirow{2}{*}{\textbf{Method}} & \multicolumn{2}{c|}{\textbf{Twitter-15}} & \multicolumn{2}{c}{\textbf{Twitter-17}} \\
                            &                         & Acc            & W-F1          & Acc            & W-F1          \\
\midrule
\multirow{2}{*}{Text}       & BiLSTM \citep{zhou2016bilstm}                  & 67.98          & 57.30         & 55.92          & 51.69         \\
                            & BERT \citep{jacob2019bert}                    & 73.87          & 70.23         & 68.48          & 65.53         \\
\midrule
\multirow{2}{*}{Image+Text} & MIMN \citep{xu2019mimn}                   & 71.84          & 65.69         & 65.88          & 62.99         \\
\multirow{2}{*}{(Original Encoders)} & TomBERT \citep{2019ijcaiTomBERT}                & 77.15          & 71.75         & 70.34          & 68.03         \\
                            & CLMLF  \citep{li2022clmlf}                 & 78.11          & 74.60         & 70.98          & 69.13         \\
\midrule
\multirow{3}{*}{Image+Text} & $\text{Concat}^\ddagger$ & 75.34          & 73.20         & 68.96          & 66.38         \\
\multirow{3}{*}{(SwinT+BERT)} & $\text{Att}^\ddagger$ & 76.01          & 73.87         & 69.54          & 67.82         \\
                            & $\text{CLMLF}^\ddagger$ \citep{li2022clmlf}                   & \underline{78.65}           & \underline{76.13}         & \underline{72.69}          & \underline{72.45}         \\
                            & CoDe (\textbf{Ours})  & \textbf{80.83}           & \textbf{79.53}         & \textbf{74.86}          & \textbf{74.36}    \\
\bottomrule
\end{tabular}}
\vskip -0.1in
\label{aspect}
\end{table}

\subsection{Generalization to Other Multimodal Datasets}
\label{generalization}
To validate the generalizability of CoDe, we further evaluate it on aspect-based multimodal sentiment datasets Twitter-15 \citep{aaai2018t15} and Twitter-17 \citep{acl2018t17}. In contrast to multimodal sentiment detection, which focuses on the overall sentiment of the post, aspect-based multimodal sentiment analysis aims to extract the sentiment of a specific aspect within the text of the post. To accommodate such variance, we add the aspect term at the end of the text when inputting the post into CoDe. We report accuracy (\textbf{Acc}) and weighted-F1 (\textbf{W-F1}) following previous methods, and present the results in \cref{aspect}. CoDe consistently outperforms both unimodal and multimodal methods, further validating its effectiveness and generalizability.

\begin{table}[t]
  \centering
    \caption{Results of incorporating in-image text into SOTA methods.}
    \vskip -0.1in
  \resizebox{1\linewidth}{!}{
  \begin{tabular}{c|l|cc|cc|l|cc|cc}
    \toprule
    \multirow{2}*{\textbf{Modality}} & \multirow{2}*{\textbf{Method}} & \multicolumn{2}{c|}{\textbf{MVSA-Sinlge}} & \multicolumn{2}{c|}{\textbf{TumEmo}} & \multirow{2}*{\textbf{Method}} & \multicolumn{2}{c|}{\textbf{MVSA-Sinlge}} & \multicolumn{2}{c}{\textbf{TumEmo}}\\
                    &    & Acc     & W-F1     & Acc     & W-F1  & & Acc     & W-F1     & Acc     & W-F1 \\              
    \midrule
    \multirow{1}*{Image} & SwinT & 64.89  & 63.94 & 49.60 & 48.27 & SwinT+In-image Text & 67.85  & 66.92 & 53.75 & 53.54 \\
    \midrule
    \multirow{1}*{Text} & BERT & 71.11  & 69.70 & 62.12 & 60.78 & BERT+In-image Text & 73.39  & 72.65 & 66.17 & 65.89 \\
    \midrule
    \multirow{5}*{\makecell[c]{Image+Text \\ (SwinT+BERT)}} & Concat                    & 73.17  & 72.41 & 70.81 & 69.97 & Concat+In-image Text                    & 73.67  & 72.99 & 71.21 & 71.20 \\
    & Att & 74.50  & 73.86 & 71.24 & 71.23 & Att+In-image Text & 74.94  & 74.25 & 71.46 & 71.43 \\
    & CLMLF & 75.61  & 74.79 & 71.30 & 71.27 & CLMLF+In-image Text & 75.61  & 74.93 & 71.72 & 71.51 \\
    & MVCN & 75.83  & 75.33 & 71.72 & 71.58 & MVCN+In-image Text & 76.05  & 75.53 & 71.98 & 71.90\\
    & - & -  & - & - & - & CoDe (\textbf{Ours}) & \textbf{77.02}  & \textbf{76.41} & \textbf{73.11} & \textbf{73.09} \\
    \bottomrule
  \end{tabular}}
    \vskip -0.1in
  \label{incorporate ocr}
\end{table}

\section{Analysis}
\subsection{Effectiveness of In-image Texts}
As pointed out in \cref{different modules}, the performance gains brought by the semantics completion module may not be strictly confined to sentiment discrepancy. Since this module introduces additional knowledge of in-image text compared to previous works, we further investigate the source of its effectiveness. Specifically, we also incorporate in-image text into the SOTA methods by fusing it into one of their modalities, and present the results in \cref{incorporate ocr}. Observably, the knowledge of in-image text only brings limited performance gains to multimodal methods, indicating that it overlaps with the knowledge of image or text in most posts, and therefore is not the main reason for our improvements. On the other hand, incorporating in-image text into unimodal methods leads to more obvious performance gains. Specifically, it brings a 2.96\% accuracy improvement for SwinT and 2.28\% accuracy improvement for BERT on MVSA-Single. This highlights the strong complementary role of in-image text when the semantic context is limited, and confirms its utility in enhancing unimodal representations. Based on these results, we speculate that the effectiveness of the semantics completion module arises from its tightening of the semantics connection between image and text by incorporating the shared in-image text knowledge into them simultaneously. In addition to alleviating potential sentiment discrepancy, it also facilitates subsequent inter-modal interactions, including both the semantics decomposition module and the global fusion of modalities.

\begin{table}[]
\centering
\caption{Computational complexity of CoDe under different unimodal encoders.}
\vskip -0.1in
\resizebox{1\linewidth}{!}{
\begin{tabular}{l|cc|cc|cc|cc}
\toprule
\multirow{2}{*}{\textbf{CoDe}} & \multicolumn{2}{c|}{Image Encoder} & \multicolumn{2}{c|}{Text Encoder} & \multicolumn{2}{c|}{Additional} & \multicolumn{2}{c}{Total} \\
                     & GFLOPs         & Params        & GFLOPs        & Params        & GFLOPs       & Params       & GFLOPs     & Params    \\
\midrule
SwinT+BERT     & 25.50          & 70.08M            & 10.88         & 85.64M            & 12.14        & 16.54M           & 48.52      & 172.26M       \\
ViT+BERT       & 16.86          & 85.64M            & 10.88         & 85.64M            & 13.18        & 16.54M           & 40.92      & 187.82M       \\
ResNet-50+BERT & 16.53          & 47.02M            & 10.88         & 85.64M            & 20.48        & 18.11M           & 47.89      & 150.77M   \\  
\bottomrule
\end{tabular}}
\vskip -0.1in
\label{complexity}
\end{table}

\subsection{Computational Complexity}
We report the computational overhead and model scale of CoDe in \cref{complexity}. Compared to the inherent overhead of unimodal encoders, CoDe only introduces comparable additional computational overhead and a relatively small number of extra parameters. This demonstrates CoDe's time efficiency and its applicability across various scenarios, including real-time or resource-limited situations.

\begin{table}[t]
  \centering
    \caption{Comparison of model's performance under different levels of perturbations in the extracted in-image text.}
    \vskip -0.1in
  \resizebox{0.8\linewidth}{!}{
  \begin{tabular}{c|cc|cc|cc|cc}
    \toprule
    \multirow{2}*{\textbf{Mask Ratio}} & \multicolumn{2}{c|}{\textbf{MVSA-Sinlge}} & \multicolumn{2}{c|}{\textbf{MVSA-Multiple}}  & \multicolumn{2}{c|}{\textbf{TumEmo}} & \multicolumn{2}{c}{\textbf{HFM}}\\
                                       & Acc     & W-F1     & Acc     & W-F1   & Acc     & W-F1     & Acc     & M-F1\\                               
    \midrule
    0\%   & \textbf{76.98}  & \textbf{76.47} & \textbf{72.74}  & \textbf{70.71}  & \textbf{73.02}  & \textbf{73.02} & \textbf{88.88}  & \textbf{88.72}  \\
    25\% & 76.79  & 76.32 & 72.40  & 69.89  & 72.90 & 72.43  &87.44 &87.41\\
    50\%  & 76.34  & 76.02 & 71.71 & 69.20  & 72.38 & 71.99 & 86.13 & 86.07 \\
    75\% & 75.56  & 75.11 & 70.88  & 69.43  & 71.32 & 70.79  &85.46 &84.99\\
    100\% & 75.39  & 74.15 & 70.74  & 69.81 & 72.49 & 71.78 &86.93 &86.91  \\
    \bottomrule
  \end{tabular}}
  \vskip -0.1in
  \label{scene text quality}
\end{table}

\subsection{Impact of In-image Text Quality}
We investigate the impact of in-image text quality on the overall performance in \cref{scene text quality}. Due to the lack of ground truth, we simulate five levels of quality by randomly replacing each word in the detected in-image text with a \textless mask\textgreater\, token, with probabilities of 0\%, 25\%, 50\%, 75\%, and 100\%. There is a continuous decline in the model's performance as the quality of in-image text deteriorates. This suggests that the lack of clarity in in-image text hinders our model's ability to alleviate the sentiment discrepancy. From another perspective, this result also highlights the potential positive impact of enhancing the quality of in-image text, revealing a direction for further improvements in future works.

\begin{figure}[t]
  \centering
  \includegraphics[width=0.8\linewidth]{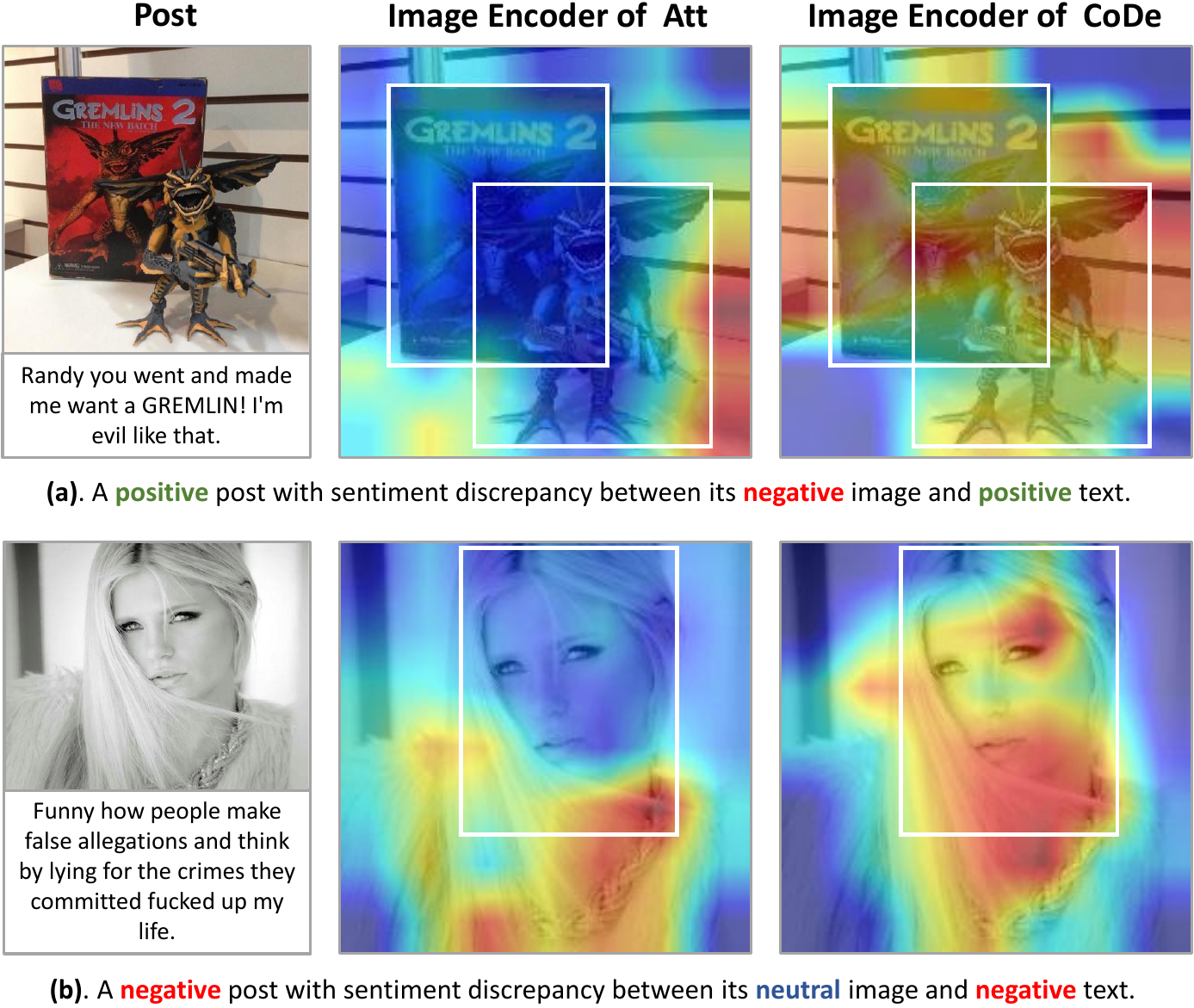}
  \vskip -0.1in
  \caption{Attention heatmaps of the image encoders of Att and CoDe for posts containing sentiment discrepancy. The white bounding boxes circle the foreground objects through which humans intuitively perceive visual sentiments.}
  \label{attention visualization}
  \vskip -0.1in
\end{figure}

\subsection{Attention Visualization}
To visually demonstrate the advantages of CoDe in resolving sentiment discrepancy compared with the single-branch fusion framework (\cref{structure} (b)), we present the attention heatmaps of their image encoders in \cref{attention visualization}. For the single-branch fusion framework, we adopt Att, which is equivalent to CoDe with the semantics completion module and semantics decomposition module removed. We show two posts containing sentiment discrepancy, with their image sentiments distinct from the overall sentiments. It can be observed that the image encoder of Att overfits toward the overall sentiments of the posts. Instead of focusing on foreground objects that reflect the correct sentiments of the images, such as the monster book and toy in the first post, the image encoder of Att mainly perceives sentiments from the backgrounds that are more aligned with the overall sentiments. This tendency leads Att to ignore certain unimodal sentiments, thereby hampering its modeling of multimodal sentiments during inference. In contrast, the image encoder of CoDe correctly concentrates on the foreground objects, sharing a perception similar to that of humans. It explains the effectiveness of CoDe from the perspective of unimodal encoding.

\begin{figure}[h]
  \centering
  \includegraphics[width=0.55\linewidth]{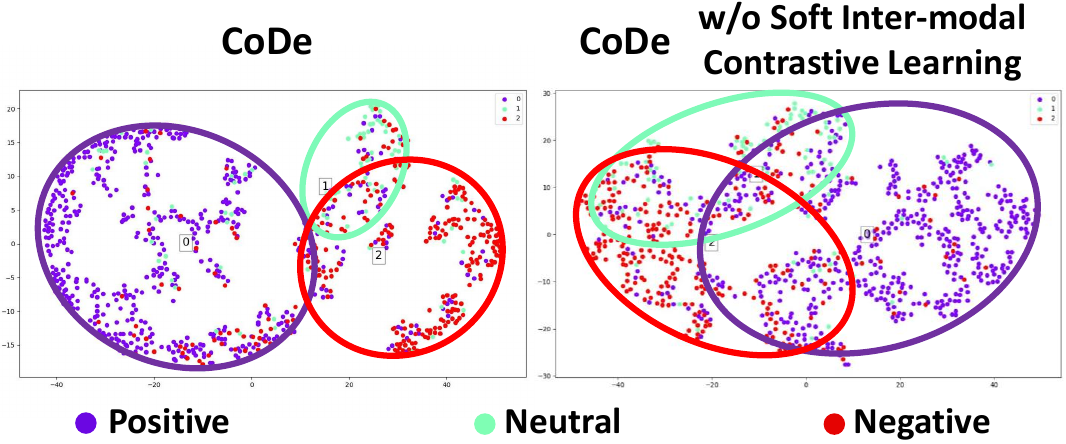}
  \vskip -0.1in
  \caption{Distribution visualization of multimodal sentiment representations on the test set of MVSA-Single with (\textbf{left}) / without (\textbf{right}) the soft inter-modal contrastive learning.}
  \label{Cluster}
  \vskip -0.1in
\end{figure}

\subsection{Distribution Visualization}
In \cref{different modules}, we verify the critical role of the soft inter-modal contrastive learning within the semantics decomposition module. A noteworthy observation reveals that removing the contrastive learning component yields inferior performance on MVSA-Single compared to complete removal of the semantics decomposition module itself. To further investigate this phenomenon, we apply t-SNE\footnote{https://github.com/mxl1990/tsne-pytorch} algorithm to visualize the feature distribution of the multimodal sentiment representation (the concatenation of $C_i, D_i$) on MVSA-Single. As demonstrated in \cref{Cluster}, the soft inter-modal contrastive effectively reduces inter-category feature overlap and enhances intra-cluster compactness. It underscores its benefits on the consistent sentiment captured by modality fusion, as it is a component of the multimodal sentiment representation. We attribute this to the feature alignment of sub-representations encouraged by the contrastive learning. Such alignment extends to the complemented representations, enabling the inter-modal cross-attention to capture representative consistent sentiments during fusion.

\begin{figure*}[h]
  \centering
  \includegraphics[width=1\textwidth]{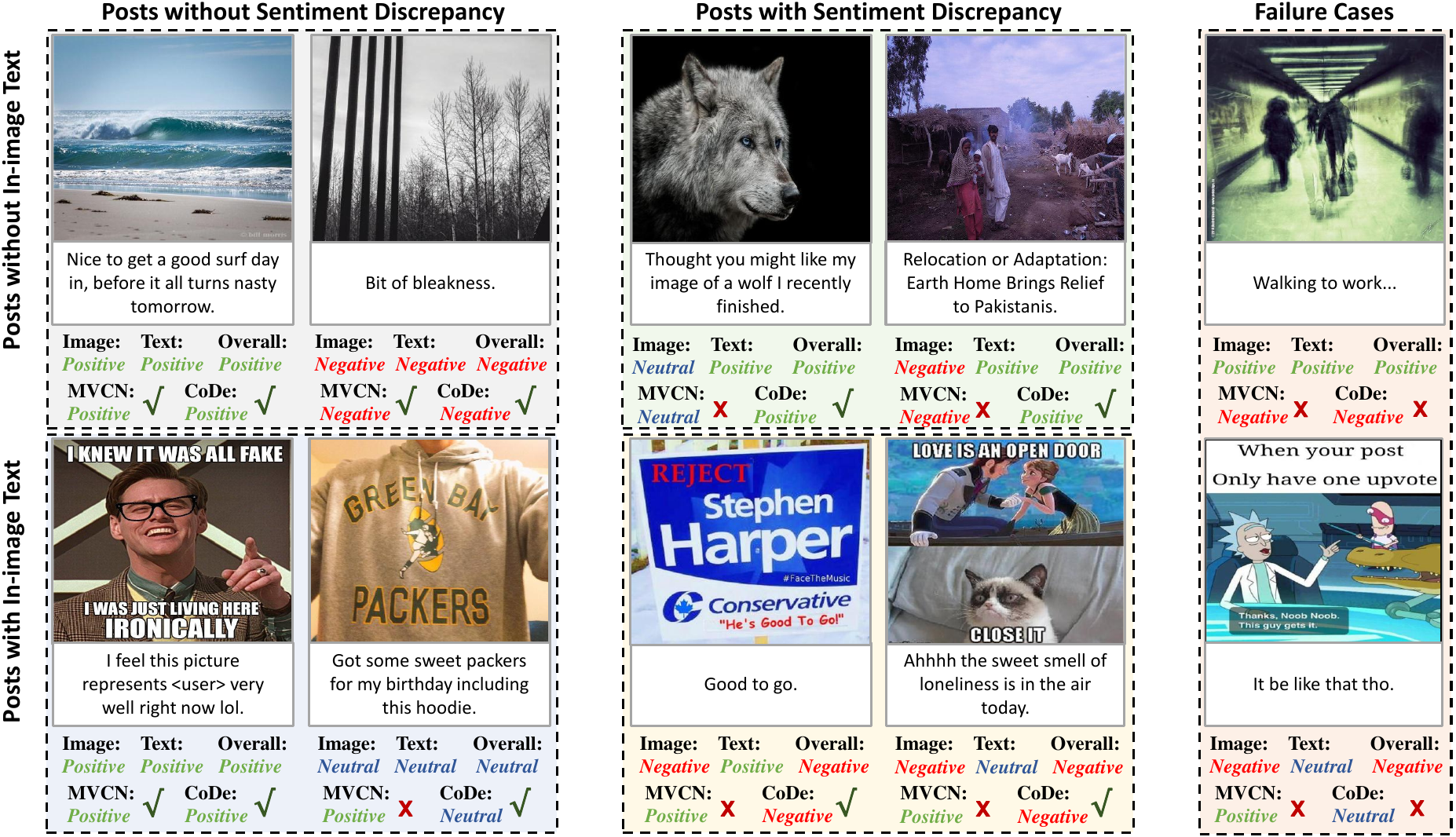}
  \vskip -0.1in
  \caption{Case study of CoDe. Posts are arranged according to whether they contain sentiment discrepancy and in-image texts. Below each post, we present the true sentiment categories of the image, the text, and the overall post in the first row; the classification results of MVCN \citep{2023aclMVCN} and CoDe in the second row.}
  \vskip -0.1in
  \label{Case study}
\end{figure*}


\subsection{Case Study}
To concretely demonstrate the effectiveness of CoDe, we conduct a comprehensive case study in \cref{Case study} comparing against MVCN \citep{2023aclMVCN}, the current state-of-the-art baseline. Our analysis reveals four key findings: \textbf{1)}. Both CoDe and MVCN correctly predict the overall sentiment for posts with neither sentiment discrepancy nor in-image text. \textbf{2)}. When posts contain in-image text, MVCN fails in the second case due to lexical traps (e.g., sweet, birthday) in the text. CoDe, on the other hand, succeeds in both posts by tightening the semantics between image and text with in-image text. \textbf{3)}. For posts with sentiment discrepancy, MVCN can hardly capture the correct sentiments of posts due to its ignorance of this issue. In contrast, CoDe accurately perceives these sentiments, reaffirming its competence in modeling discrepant sentiment. \textbf{4)}. Under the combined challenges of sentiment discrepancy and in-image text, CoDe still makes reliable predictions by complementing the semantics of in-image text with those of the image and text and explicitly modeling the discrepant sentiment. For example, in the first case, CoDe leverages in-image texts (e.g., Stephen Harper) as contextual semantics to understand why sentiment discrepancies exist and what sentiment the post expresses.

Two representative misclassified posts are displayed on the far right of \cref{Case study}. In the first post, the image is an artwork with a unique style that the model is unfamiliar with. This leads to the model mistakenly categorizing it as negative based on the color tone. In the second post, the image depicts the dismal situation of a scientist in an animation, which the model does not correctly capture. We attribute these failures to our model's lack of robust understanding of images of different styles and the necessary external knowledge. Therefore, tackling these problems may promote further improvements in multimodal sentiment detection in the future.

\section{Conclusion and Discussion}
In this paper, we propose CoDe to resolve sentiment discrepancy in multimodal sentiment detection. It contains two modules. The semantics completion module complements image and text representations with the semantics of in-image text, and the semantics decomposition module decomposes the modality-shared and modality-private sentiments within the unimodal representations. Benefiting from these modules, CoDe explicitly captures discrepant sentiments and integrates them with consistent sentiments from modality fusion, ultimately forming comprehensive multimodal representations. Systematic evaluations across four datasets demonstrate CoDe's superiority and confirm the necessity of resolving sentiment ambiguity. Comprehensive ablation studies verify the semantics completion module's effectiveness in complementing unimodal semantics and bridging cross-modal sentiment gaps, as well as the semantics decomposition module's capability to align cross-modal features while decoupling discrepant sentiments.

A primary limitation of this paper lies in CoDe's potential sensitivity to in-image text quality. Since ground truth annotations are unavailable, in-image text must be extracted through external tools, and noise from this process may adversely affect model performance. As a popular research field, multimodal sentiment detection still presents several avenues for improvement. Potentially promising directions include enhancing model robustness, incorporating external knowledge, and addressing sentiment ambiguity arising from subjective perception.

\section*{Acknowledgment}
This work is supported by the National Natural Science Foundation of China (Grant No. 62376266 \& 62406318).

\bibliography{main}
\bibliographystyle{iclr2026_conference}

\end{document}